\documentclass[lettersize,journal]{IEEEtran}
\usepackage[
  colorlinks=true,
  linkcolor=magenta,
  urlcolor=magenta
]{hyperref}
\usepackage{amsmath,amsfonts}
\usepackage{algorithmic}
\usepackage{algorithm}
\usepackage{array}
\usepackage[caption=false,font=normalsize,labelfont=sf,textfont=sf]{subfig}
\usepackage{textcomp}
\usepackage{stfloats}
\usepackage{url}
\usepackage{verbatim}
\usepackage{graphicx}
\usepackage{cite}
\usepackage{multirow}
\usepackage{booktabs}
\usepackage{siunitx}
\usepackage{xcolor}
\usepackage[switch, modulo]{lineno}
\begin{document}

\title{Pyramid-Monozone Synergistic Grasping Policy in Dense Clutter}
\author{Chenghao Li, \IEEEmembership{Student Member, IEEE}, and Nak Young Chong, \IEEEmembership{Senior Member, IEEE}
\thanks{This work was supported by JSPS KAKENHI Grant Number JP23K03756 and the Asian Office of Aerospace Research and Development under Grant/Cooperative Agreement Award No. FA2386-22-1-4042.}
\thanks{The authors are with the School of Information Science, Japan Advanced Institute of Science and Technology, Ishikawa 923-1292 Japan (e-mail: chenghao.li@jaist.ac.jp; nakyoung@jaist.ac.jp).}}

\markboth{Journal of \LaTeX\ Class Files,~Vol.~14, No.~8, August~2021}%
{Shell \MakeLowercase{\textit{et al.}}: A Sample Article Using IEEEtran.cls for IEEE Journals}


\maketitle

\begin{abstract}

Grasping a diverse range of novel objects in dense clutter poses a great challenge to robotic automation mainly due to the occlusion problem. In this work, we propose the Pyramid-Monozone Synergistic Grasping Policy (PMSGP) that enables robots to effectively handle occlusions during grasping. Specifically, we initially construct the Pyramid Sequencing Policy (PSP) to sequence each object in cluttered scenes into a pyramid structure. By isolating objects layer-by-layer, the grasp detection model is allowed to focus on a single layer during each grasp. Then, we devise the Monozone Sampling Policy (MSP) to sample the grasp candidates in the top layer. Through this manner, each grasp targets the topmost object, thereby effectively avoiding most occlusions. We performed more than 7,000 real-world grasping in densely cluttered scenes with 300 novel objects, demonstrating that PMSGP significantly outperforms seven competitive grasping methods. More importantly, we tested the grasping performance of PMSGP in extremely cluttered scenes involving 100 different household goods, and found that PMSGP pushed the grasp success rate to 84.9\%. To the best of our knowledge, no previous work has demonstrated similar performance. All grasping videos are available at: \href{https://www.youtube.com/@chenghaoli4532/playlists}{https://www.youtube.com/@chenghaoli4532/playlists}.
\end{abstract}

\begin{IEEEkeywords}
Robot grasping, grasp detection, deep learning, dense clutter, grasp sequence policy, grasp sampling.
\end{IEEEkeywords}

\section{INTRODUCTION}

\begin{figure}[!t]
\vspace{0.2\baselineskip}
\centerline{\includegraphics[width=\columnwidth]{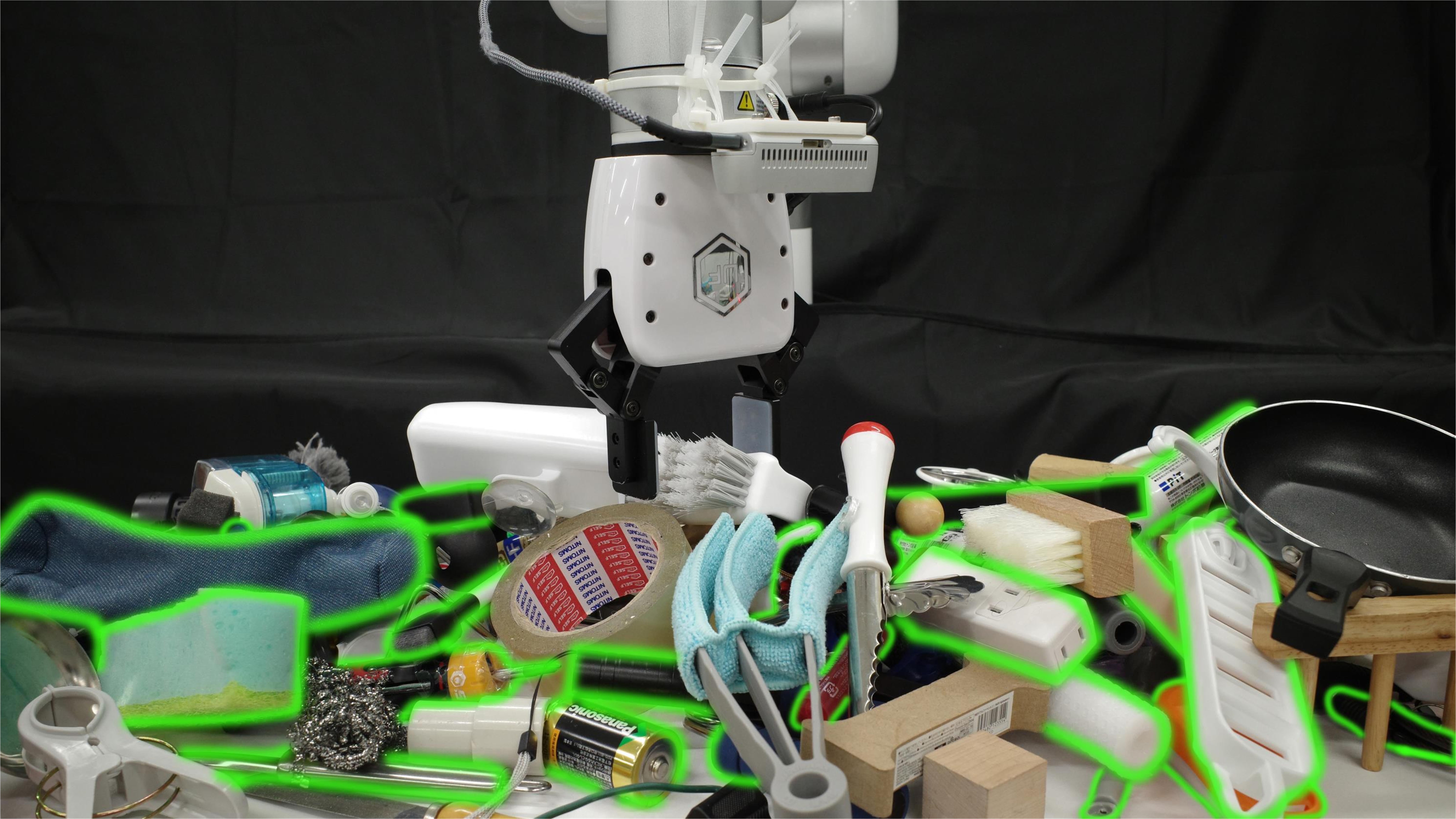}}
\caption{Grasping in dense clutter: Objects are occluded to one another (highlighted in green lines) and partially visible, which can easily lead to failed grasping.}
\label{fig1}
\end{figure}

\IEEEPARstart{G}{rasping} a wide range of novel objects can benefit applications such as warehousing, manufacturing, retail, and services. However, when novel objects are densely cluttered, occlusions can make this task extremely challenging. As shown in  Fig. \ref{fig1}, incomplete object shapes due to occlusions may greatly undermine grasping capability.

One approach to dense clutter grasping is to build a database of grasps on three-dimensional (3D) object models, using performance metrics based on geometry and physics \cite{c1, c2}, along with stochastic sampling for uncertainty quantification \cite{c3}. Although these techniques provide robust solutions in structured environments, they are fundamentally constrained by their dependence on complete 3D object models. In real-world applications, such models are not always accessible, especially when robots operate in unstructured environments with numerous unknown objects. Additionally, building accurate 3D models is often labor-intensive, expensive, and time-consuming, demanding significant manual effort for each unique object. These limitations underscore the need for more adaptable and efficient approaches to robot grasping that can cope with uncertainty and variations in object geometry. In response to these challenges, there has been a new approach \cite{c4, c5} leveraging deep neural networks (DNN) \cite{c6, c7, c8, c9, c10} to train the function approximator. The approximator predicts the best grasp candidate from images based on quality score-based sampling utilizing large datasets of empirical successes and failures. This approach can generalize to novel objects at a significantly low cost. However, the quality score-based sampling used in these methods rarely considers the occlusion problem, which potentially leads the optimal grasp predicted on occluded objects to a collision.

A common solution to the problem mentioned above is to first segment all objects in the scene to create a mask, and then use this mask to guide the sampling of the quality score \cite{c11, c12}. This involves evaluating the relation of each instance (object) and determining whether their grasp candidates might result in collisions. But, this method is constrained by the quality score itself, that is, the quality score does not guarantee that every instance has a viable grasp candidate, even though the instance is not occluded. Furthermore, this method typically requires classifying and segmenting all objects in the scene, and these segmented objects may also have incomplete shapes due to occlusion. Thus, this method still does not effectively solve the occlusion problem and essentially trades high costs for limited improvements.

Now we look at the grasp problem in clutter from a different perspective. Since the robot typically grasps one object at a time, why not identify an unoccluded region before each grasp attempt and only sample within this region, which would not only be accurate but also avoid redundant analysis? 

In this paper, we present a novel grasping policy, called the Pyramid-Monozone Synergistic Grasping Policy (PMSGP), which leverages the Pyramid Sequencing Policy (PSP) to sequence each object in the scene into a pyramid structure, meaning that before each grasp, only the top layer instance is segmented. This allows the grasp detection model to focus on the top layer. Then, through the Monozone Sampling Policy (MSP), the grasp sampling is also concentrated on the top-layer instance. More precisely, we do not use the mask to guide the quality score-based sampling, but directly sample these grasp candidates within the mask. This pyramid and monozone operation help the robot avoid most occlusions during grasping, and as the grasping progresses, the extent of occlusion gradually decreases, thereby reducing the difficulty of subsequent grasps. 
Our contributions are as follows:
\begin{enumerate}
\item We propose a novel dense clutter grasping policy incorporating pyramid sequencing and monozone sampling policies.
\item We restructure the problem of grasping novel objects in dense clutter into a hierarchically layered grasp detection problem, each layer of which contains only a single object. This allows sampling grasp candidates from each single layer.
\item We conduct extensive real-world grasping experiments, and demonstrate that our method far outperforms seven competitive methods among 300 novel objects in various densely cluttered scenes. More importantly, even in extremely cluttered scenes with up to 100 household goods, our method still exhibits a high grasp success rate, reaching 84.9\%.
\end{enumerate}

This paper is organized into the following sections. Section II (Related Work) provides a review of traditional grasping methods and learning-based grasping methods. Section III (Grasp Configuration) describes the 4-DOF grasp configuration, and how to transform it from the image coordinates to the robot end effector coordinates. Section IV (Proposed Method) provides an overview of PMSGP and makes a detailed description of its two components (PSP and MSP), as well as each submodule of each component. Section V  (Experiments) shows the real grasping results of our method and two groups of baseline methods in dense clutter and performed a failure analysis. Finally, Section VI (Conclusion) summarizes the work of this paper and provides prospects for future research.

\section{Related Work}
While many grasping frameworks exist, this work only focuses on vision-guided 4-DOF grasping with a parallel-jaw gripper. The 4-DOF grasp framework typically performs grasping in a top-down manner, where the robot moves along the $X$, $Y$, and $Z$-axis and rotates only around the $Z$-axis. During grasping, the parallel-jaw gripper will adjust its opening stroke based on the size of the object perceived by the depth camera. It is mainly divided into traditional methods and learning-based methods as follows.

\subsection{Traditional Grasping Methods}
Traditional grasping methods rely on mathematical and physical models that describe the geometry, kinematics, and dynamics of objects \cite{c1, c2, c3}. These methods typically assume access to a detailed 3D model of the object, which is used to compute stable grasps. For example, Gallegos {\it et al.} \cite{c13} optimized grasp strategies by leveraging both a known 3D model of the object and predefined contact points for the robot gripper. Similarly, Pokorny {\it et al.} \cite{c14}  proposed grasping spaces, where objects could be mapped to these spaces to identify suitable grasps. While these techniques offer robust solutions in controllable structured environments, they are inherently limited by their reliance on complete 3D object models. In real-world scenarios, such models may not always be available, particularly when robots are deployed in uncontrollable unstructured environments with many unknown objects. Moreover, the process of constructing accurate 3D models is often labor-intensive, costly, and time-consuming, requiring extensive manual effort for each unique object. Therefore, these constraints highlight the need for more adaptable and efficient approaches to robot grasping that can handle uncertainty and variability in object geometry.

\subsection{Learning-based Grasping Methods}
Learning-based methods can generalize to various novel objects in dense clutter, which typically involve training a function approximator, such as DNN, to predict the success probability of grasp candidates from images by leveraging large datasets of empirical successes and failures. For that reason, datasets play a crucial role in these methods. One human-labeled dataset is the Cornell Grasping Dataset \cite{c15}, which contains around 1,000 RGB-D images and has been widely used to train grasping models, such as \cite{c17, c18, c19, c20, c21, c22, c23}, based on convolutional neural networks (CNN) \cite{c16}. However, this dataset is quite small and consists only of single-object images, which limits the dense clutter grasping capabilities of models trained on it.

The Dex-Net series \cite{c4, c24, c25, c26, c27} made significant advancements by generating large synthetic datasets that incorporate various dense clutter scenes. Despite these advancements, this approach did not fully resolve the sim-to-real problem. GraspNet \cite{c5, c28, c29}, in contrast, constructed a real-world dataset featuring one billion grasp labels and nearly 100,000 images with 190 different dense clutter scenes, supporting both 4-DOF and 6-DOF grasping. This dataset enabled remarkable real grasping performance in dense clutter. However, the sampling used in the above methods relies on predicting the success probability (quality score) of the grasp candidates, which almost does not consider the occlusion problem, potentially leading the optimal grasp predicted on occluded objects to a collision.

Recently, several works proposed to segment all objects in a scene to create a mask that can guide the sampling of quality scores \cite{c11, c12}. These works involve evaluating the relationships between objects and assessing whether each grasp candidate might result in collisions. However, the effectiveness of this approach is also limited by the quality score itself. Specifically, the quality score does not ensure that every object has a viable grasp candidate, even when the object is not occluded. Additionally, it typically requires classifying and segmenting all objects in the scene, and these segmented objects may still have incomplete shapes due to occlusions. Therefore, this approach also fails to effectively address the occlusion issue and essentially exchanges high costs for only marginal improvements. Unlike the above works, we first organize each object in the scene into a pyramid structure, segmenting only the topmost object during each grasp. Segmenting in a layer-by-layer manner, most grasp candidates can be restricted to a single layer. We then sample grasp candidates for the object in the top layer for each grasp. This allows the robot to avoid most occlusions during grasping.

\begin{figure*}[!t]
\centerline{\includegraphics[width=\textwidth]{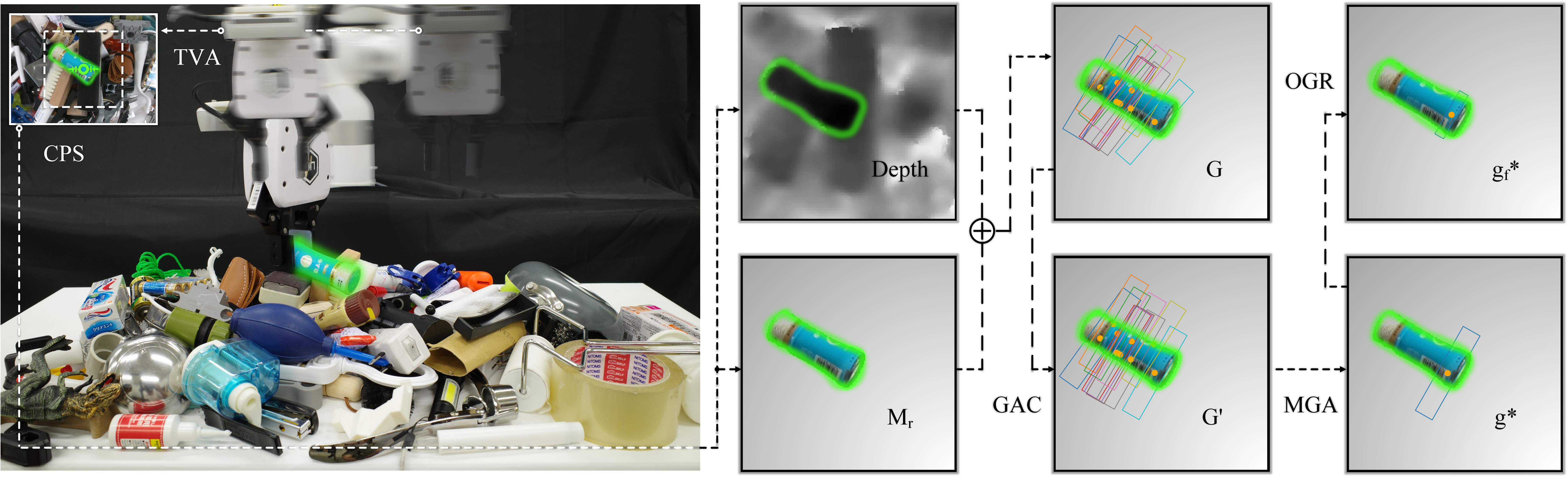}}
\caption{Pipeline of PMSGP: Firstly, conducting Top View Alignment (TVA) to center the initial view $V$ of depth camera on the topmost object to get view $V'''$, and segment this topmost object by the center $V'''(c)$ (green point) of this view as prompt to obtain initial segmented RGB image (emphasized with green lines) with mask $M_f$. Then, calculate two pairs of most distant points ($p_m$ (red point), $p_m'$ (red point), ${p_{m_p}}$ (blue point), and ${p_{m_p}'}$ (blue point)) based on the edge of $M_f$, and using these points to make Cross-prompted Segmentation (CPS) to optimize $M_f$ to get ${M_r}$. In step three, the segmented RGB image with mask ${M_r}$ and the depth image within view $V'''$ are fed into the grasp detection (GD) model to generate initial grasp candidates $G$, followed by Grasp Angle Calibration (GAC) to obtain calibrated grasp candidates $G'$. After GAC, $G'$ will be through Monozone Grasp Analysis (MGA) to find the optimal grasp $g^*$. Finally, $g^*$ is optimized by Optimal Grasp Refinement (OGR) to transfer it to the final grasp $g^*_f$.}
\label{fig2}
\end{figure*}

\section{Grasp Configuration}
Now, we elaborate on how the 4-DOF grasp configuration is represented in the image coordinate system and its conversion to the robot end effector coordinate system (eye-in-hand grasping). Specifically, we adopt the same grasp configuration in \cite{c18}, which is composed of parameters $(x, y, w, h, \theta)$ forming a rotated box. Here, $(x, y)$ represents the center of the box, $w$ and $h$ denote the width and height of the box, and $\theta$ represents the angle of the box relative to the horizontal direction. 

Since $h$ is used only for visual representation and not in the conversion process, we denote the grasp configurations in the image and robot end effector coordinate systems as $g_{i} (x, y, w, \theta)$ and $g_{r} (x_r, y_r, z_r, w_r, \theta_r)$, respectively. Here, $(x_r, y_r, z_r)$ represents the grasp position in the robot end effector coordinate system, $w_r$ is the opening stroke of the parallel jaw gripper, and $\theta_r$ is the rotation angle of the gripper relative to the $Z$ axis. The parameter conversion between $g_{i}$ and $g_{r}$ is described in Eq. \ref{eq:1}, Eq. \ref{eq:2}, and Eq. \ref{eq:3}.

The conversion process is divided into two parts. The first part involves converting $(x, y)$: using depth information ($d$) and the camera's intrinsic parameters ($f_x$, $f_y$ for focal lengths and $c_x$, $c_y$ for the image center coordinates), we convert $(x, y)$ from the image coordinate system to the camera coordinate system $(x_c, y_c, z_c)$. This is followed by converting $(x_c, y_c, z_c)$ to the robot end effector coordinate system $(x_r, y_r, z_r)$ using the positional transformation relationship $T_{rc}$. The second part involves converting $(w, \theta)$ into $(w_r, \theta_r)$ using the projection function $P_{ri}$.

\begin{equation} \label{eq:1}
\begin{bmatrix}
x_c \\
y_c \\
z_c
\end{bmatrix} = 
\begin{bmatrix}
f_x^{-1} & 0 & -c_x f_x^{-1} \\
0 & f_y^{-1} & -c_y f_y^{-1} \\
0 & 0 & 1
\end{bmatrix} 
\begin{bmatrix}
x \\
y \\
1
\end{bmatrix}  {d}
\end{equation}

\begin{equation} \label{eq:2}
\begin{aligned}
(x_r, y_r, z_r) = T_{rc}(x_c, y_c, z_c)
\end{aligned}
\end{equation}

\begin{equation} \label{eq:3}
\begin{aligned}
(w_r, \theta_r) = P_{ri}(w, \theta)
\end{aligned}
\end{equation}

The intrinsic parameters and depth information are directly obtained from the depth camera, and $T_{rc}$ is derived from offline eye-in-hand calibration. $T_{rc}$ depends on the relative orientations (excluding rotation) of the $X$, $Y$, and $Z$-axis between the camera coordinate system and the robot end effector coordinate system. Finally, the projection function $P_{ri}$ allows for manual adjustment of the linear relationship between the gripper stroke $w_r$ and rotation $\theta_r$ relative to the grasp box's width $w$ and rotation $\theta$.

Once the final grasp pose in the robot end effector coordinate system $(x_r, y_r, z_r, \theta_r, \theta_{x}^{*}, \theta_{y}^{*})$  is obtained, where $\theta_{x}^{*}$ and $\theta_{y}^{*}$ represent the constant rotations relative to the $X$-axis and the $Y$-axis, the gripper is moved to the target pose using inverse kinematics and its stroke is kept to the width $w_r$.

\section{Proposed Method}
We propose a novel grasping policy, the Pyramid-Monozone Synergistic Grasping Policy (PMSGP), designed to improve the grasping performance of parallel-jaw grippers in dense clutter as illustrated in Fig. \ref{fig2}. PMSGP is composed of two main modules: the Pyramid Sequencing Policy (PSP) and the Monozone Sampling Policy (MSP). The PSP is used for layering objects in cluttered scenes, with submodules for Top View Alignment (TVA) and Cross-prompted Segmentation (CPS). The MSP samples the topmost objects based on the layering results from the PSP, which includes Grasp Angle Calibration (GAC), Monozone Grasp Analysis (MGA), and Optimal Grasp Refinement (OGR). Overall, the MSP is inherently dependent on the PSP, and the submodules within both are linked to each other. A detailed explanation of each module is given below.

\subsection{Pyramid Sequencing Policy (PSP)}
It is important to know how to determine an appropriate grasp sequence in dense clutter to decrease the difficulty of grasping. In this context, we introduce the Pyramid Sequencing Policy (PSP), organizing objects in a cluttered scene into a pyramid structure, each layer of which contains only a single object, allowing the grasp detection model to focus on the object at the top layer each time. This is similar to reversing a sequence of piling up building blocks, that is, pulling out blocks from the top layer. 

\subsubsection{Top View Alignment}
We first define the robot’s workspace (the robot base coordinate) with the $X$ and $Y$ axis limited by the edges of a 120 $cm$ $\times$ 80 $cm$ table. The range of the $Z$-axis is limited by the maximum distance to the tabletop (40 $cm$) and the minimum distance (10 $cm$) to prevent collisions between the gripper fingers and the table. The camera is mounted at the distal end of the robot arm to which the gripper (about 10 $cm$) is also removably mounted. Before each grasping attempt, we set the robot to a pre-specified position (40 $cm$ above the center of the table) and ensure that the camera covers the entire pile of objects on the table. Then, we conduct the global view alignment based on the initial depth image $V$ and its center $V(c)$ of the camera. Specifically, we align the pixel corresponding to the minimum depth value (among 1280 $\times$ 720 pixels) in $V$ with $V(c)$ by moving the camera mounted on the robot. We denote the depth image after the camera movement by $V'$. In theory, a one-time global view alignment should suffice. However, in practice, due to errors introduced by the depth camera (both intrinsic errors and those caused by exposure), the pixel location corresponding to the minimum depth value used for the aforementioned global view alignment may differ from the true location. 

\begin{figure}[!t]
\vspace{0.15\baselineskip}
\centerline{\includegraphics[width=\columnwidth]{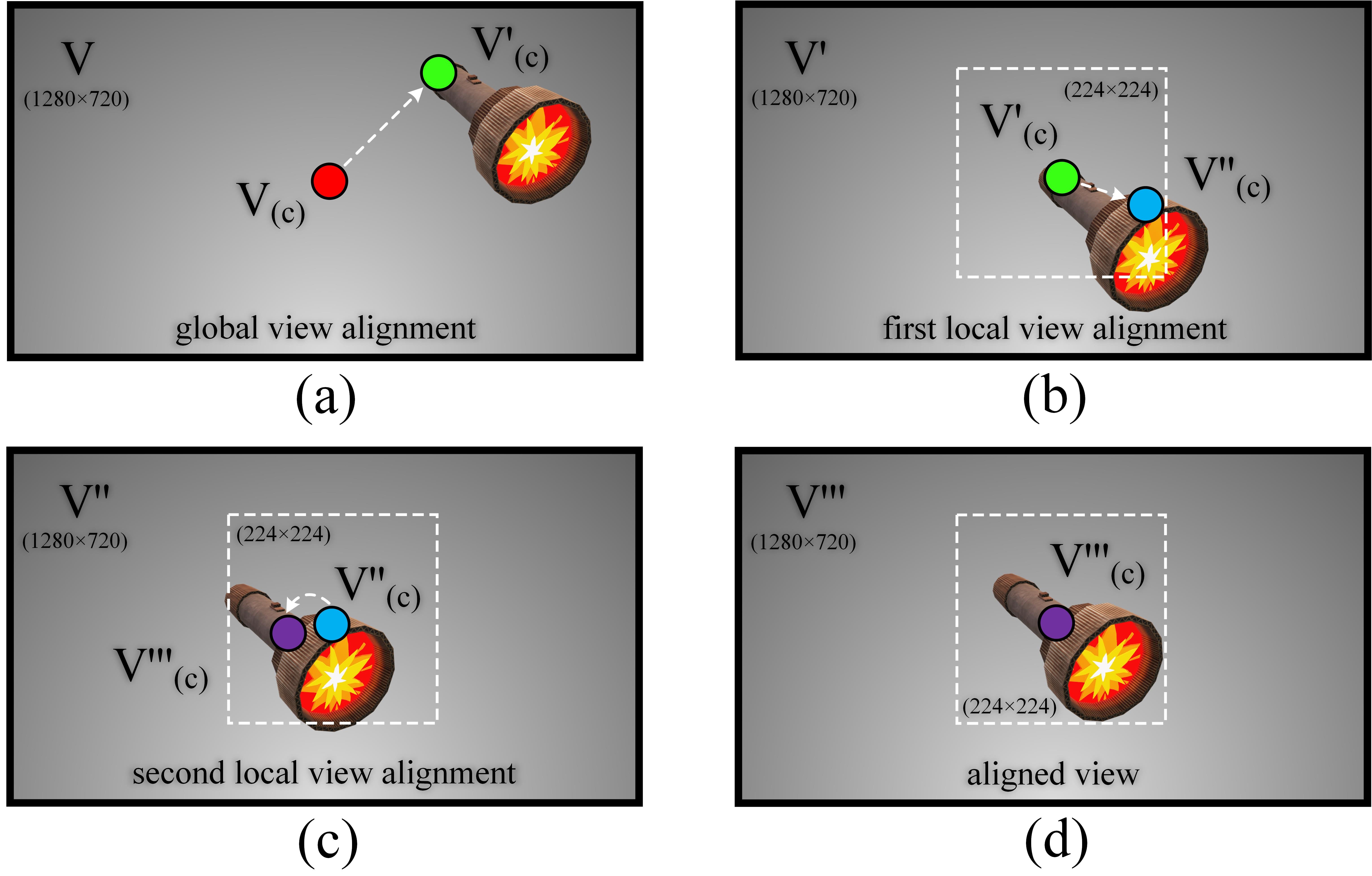}}
\caption{Sequence of top view alignment: (a) global view alignment, (b) first local view alignment, (c) second local view alignment, and (d) final aligned view. $V(c)$, $V'(c)$, $V''(c)$, and $V'''(c)$ are denoted by red point, green point, blue point, and purple point.}
\label{fig3}
\end{figure}

Therefore, we narrow the region of interest and perform local view alignment using the minimum depth value pixel in the region to mitigate the impact of depth errors and ensure that the aligned object sits also on the top of the regional pile. Specifically, to improve efficiency, we use the central block with 224 $\times$ 224 pixels of the original 1280 $\times$ 720 image to perform two consecutive local alignments, in the same way as in the global view alignment. Let $V''$ and $V'''$ denote the depth images after each local alignment operation, respectively. One reason for choosing the 224 $\times$ 224 image block is that grasp detection models typically accept images of this size as input. The sequence of alignment is illustrated in Fig. \ref{fig3}. We assume that all minimum depth value pixels stay in the same object with the camera position alignments. Finally, we use the Segment Anything Model (SAM) \cite{c30} to segment the $V'''$ to obtain the segmented RGB image with mask $M_f$. The segmentation extracts the object image that contains the pixel of the minimum depth value. Intuitively, the center pixel $V'''(c)$ of $V'''$ serves as a prompt for the segmentation.

\begin{figure}[!t]
\centerline{\includegraphics[width=\columnwidth]{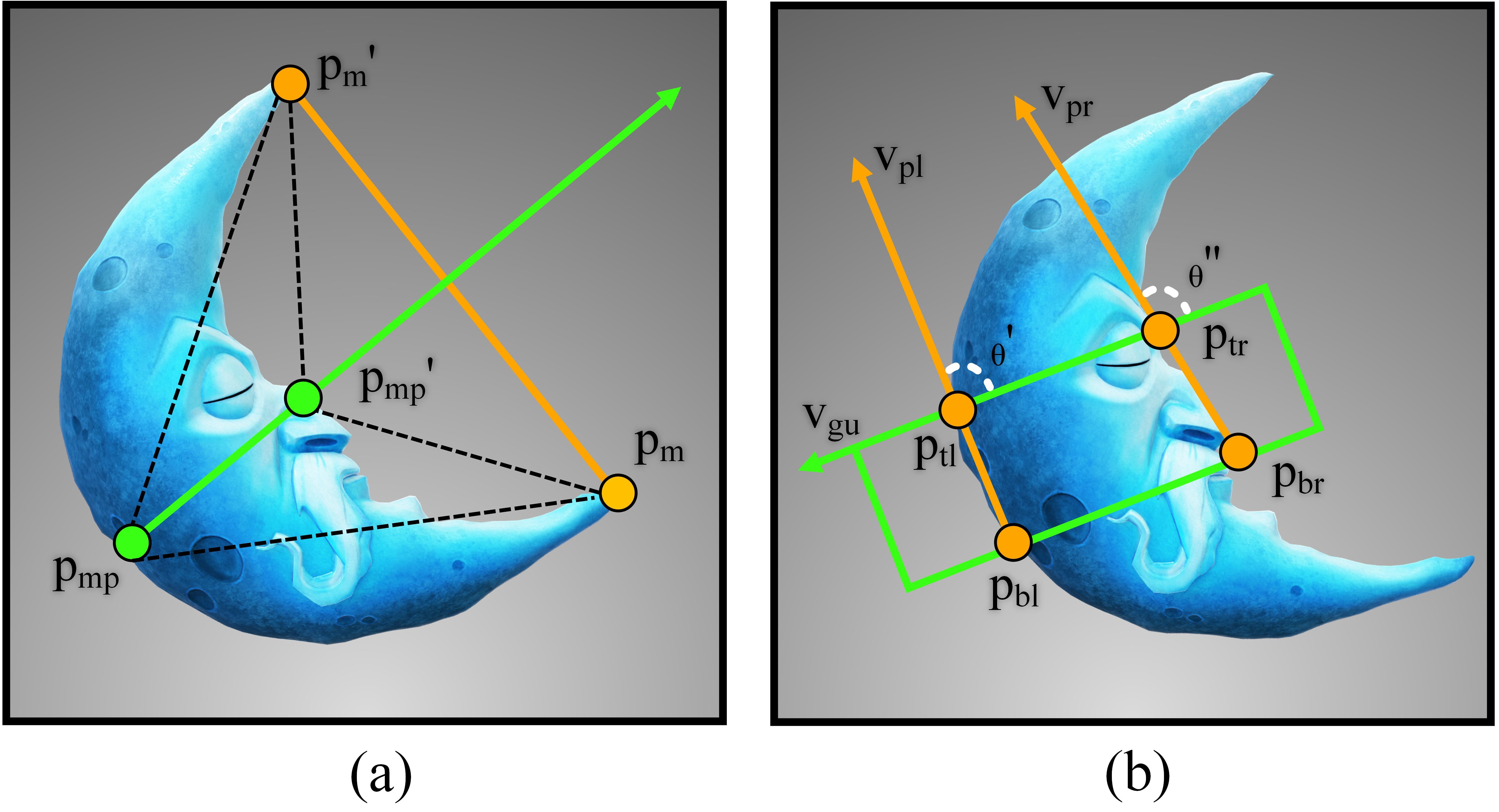}}
\caption{Visualization of (a) cross-prompted segmentation and (b) grasp angle calibration. In (a), $p_{m_p}$ and $p_{m_p}'$ are denoted by orange points, $p_{m}$ and $p_{m}'$ by green point. In (b), $p_{t_l}$, $p_{b_l}$, $p_{t_r}$, and $p_{b_r}$ by orange points, $\mathbf{v}_{p_l}$ and $\mathbf{v}_{p_r}$ by orange lines, $\mathbf{v}_{g_u}$ by green line, and  $\theta'$ and $\theta''$ by white arc.}
\label{fig4}
\end{figure}

\begin{algorithm}[H]
\caption{PSP}
\begin{algorithmic}[1]
    \STATE \textbf{Input:} Initial view $V$
    \STATE \textbf{Output:} Refined mask $M_r$\\
    // Top View Alignment: First conduct global view alignment to get to ${V'}$, then conduct two local view alignments to get ${V''}$ and ${V'''}$, and finally segment the topmost object in ${V'''}$ to get ${M_f}$.\\
    \STATE ${V'} \gets V$, $V'' \gets V'$, $V''' \gets V''$\\
    \STATE ${M_f} \gets SAM(V'''(c))$\\
    // Cross-prompted Segmentation: First find four points, then use them to perform segmentation to generate ${M_s}$, and finally refine ${M_s}$ to obtain ${M_r}$ based on ${M_d}$.\\
    \FOR{$(p_i, p_j) \in P_e \times P_e, i \neq j$}
        \STATE $(p_m, p_m') \gets \underset{\substack{(p_i, p_j) \in P_e \times P_e \\ i \neq j}}{\arg\max} \sqrt{(x_i - x_j)^2 + (y_i - y_j)^2}$
    \ENDFOR
    \STATE $(p_{m_p}, p_{m_p}') \gets \perp_{(p_m, p_m')} \cap {P_e}$\\
    \STATE ${M_s} \gets SAM(p_m, p'_m, p_{m_p}, p_{m_p}')$\\
    \STATE Perform dilation to get ${M_d}$
    \STATE ${M_r} \gets {M_s}+{M_d}$
    \RETURN ${M_r}$
\end{algorithmic}
\end{algorithm}

Through the aforementioned steps, PSP not only effectively segments the topmost object in a densely cluttered scene, which consistently appears near the center of the camera view, but also mitigates the issue of partially visible objects caused by input size constraints of the grasp detection model. It is important to note that this method is not a multi-view grasping \cite{c18}, which is a complex optimization process. In contrast, our method is more efficient given a single view depth image.

\subsubsection{Cross-prompted Segmentation}
We propose the cross-prompt to optimize segmentation in PSP because the single-point prompt is highly unstable in dense clutter. This issue is particularly pronounced when the object's appearance is complex, such as the food packaging, where only part of the object is segmented (usually manifested as many holes in the segmented object). This limitation adversely impacts the subsequent prediction and sampling of grasp candidates. 

Specifically, it begins by applying the Sobel operator \cite{c31} to extract the edges of the instance mask $M_f$ obtained from the initial single-point prompt segmentation. We then search for the two pixels most distant from each other, which we refer to as ${p_m}$ and ${p_m'}$ on the edges. As shown in Eq. \ref{eq:4}, $P_e$ means the set of all pixels on the edges, and ${P_e \times P_e}$ represents the Cartesian product of the set $P_e$. Next, we calculate the perpendicular line $\perp_{(p_m, p_m')}$ connecting ${p_m}$ and ${p_m'}$, and intersecting this perpendicular line with the edges yields another pair of the most distant pixels, which we refer to as ${p_{m_p}}$ and ${p_{m_p}'}$ as shown in Eq. \ref{eq:5} and Fig. \ref{fig4} (a). These four points are then used as prompts to perform the second segmentation, resulting in ${M_s}$. Compared to other points, these two farthest pairs of points can better exploit the geometric constraints of the initial segmentation result, thus recovering the holes in initial segmentation, and then achieving significant results in the second segmentation, as demonstrated in our ablation studies. Finally, ${M_s}$ is refined by image dilation processing: a depth threshold is applied and pixels from the first prompt serve as initial points for segmentation to produce ${M_d}$. By combining ${M_d}$ and ${M_s}$, we obtain the refined ${M_r}$ as shown in Algorithm 1.

\begin{equation} \label{eq:4}
(p_m, p_m') = \underset{\substack{(p_i, p_j) \in P_e \times P_e \\ i \neq j}}{\arg\max} \sqrt{(x_i - x_j)^2 + (y_i - y_j)^2}
\end{equation}

\begin{equation} \label{eq:5}
(p_{m_p}, p_{m_p}') = \perp_{(p_m, p_m')} \cap {P_e}
\end{equation}

\subsection{Monozone Sampling Policy (MSP)}
After isolating the topmost object by PSP, we sample the grasp candidates within this object. Here, we discard quality score-based sampling and propose the Monozone Sampling Policy (MSP) to sample grasp candidates for this object. In other words, we do not use the instance mask to guide the quality score-based sampling, but instead focus directly on sampling these grasp candidates within the mask. Moreover, MSP comprehensively considers all parameters of the grasp candidates (rotation $\theta$, position $(x, y)$, and width $w$). It analyzes the relationship between the grasp candidates and the object being grasped and its neighboring objects. Next, we elaborate on these components.

\begin{figure}[!t]
\vspace{0.2\baselineskip}
\centerline{\includegraphics[width=\columnwidth]{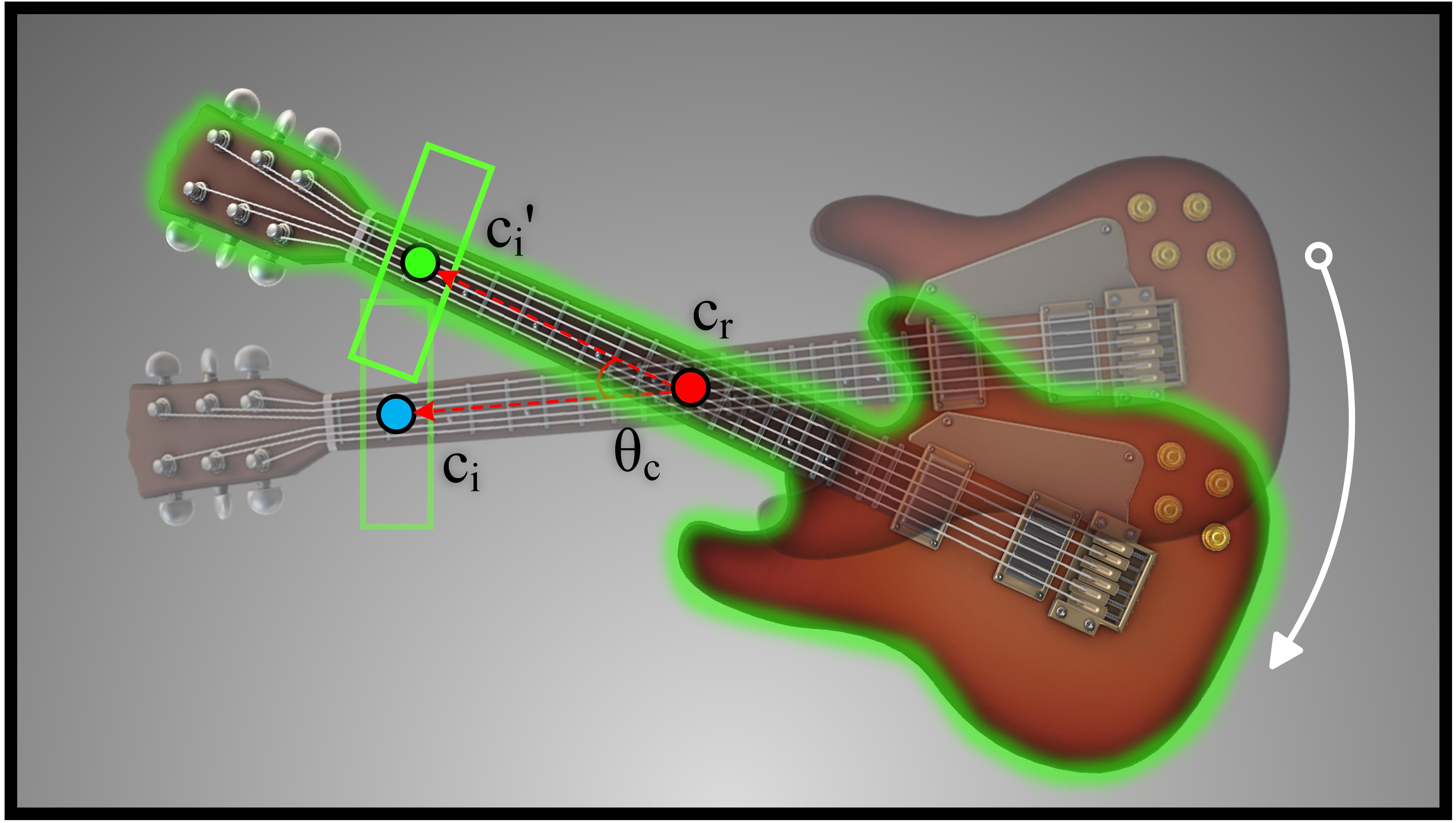}}
\caption{Visualization of adaptive viewpoint rotation in MGA. The red point, green point, and blue point are rotation center $c_r$, the center $c_i'$ of $g_i'$, and the center $c_i$ of $g_i$, respectively. The rotation angle is denoted as $\theta_c$.}
\label{fig5}
\end{figure}

\begin{algorithm}[!t]
\caption{MSP}
\begin{algorithmic}[1]
    \STATE \textbf{Input:} Initial grasp candidate sets $G$
    \STATE \textbf{Output:} Final grasp $g^*_f$\\
    // Grasp Angle Calibration: Rotate all grasp candidates $g_i$ to optimal angle $R^*$.\\
    \FOR{$g_i \in G$}
        \STATE ${R}^* \gets \underset{R} {\arg\min} \left( \left|\theta'({R}) - \frac{\pi}{2}\right| + \left|\theta''({R}) - \frac{\pi}{2}\right| \right)$ \\ 
        \STATE ${G'} \gets (G, R^*)$
    \ENDFOR\\
    // Monozone Grasp Analysis: Filter out $g_i$ that leads the robot to collide with this object itself, this object's adjacent objects, or the table to obtain the optimal grasp $g^*$.\\
    \FOR{$g_i \in G'$}
        \STATE $G'' \gets \{ g_i \in G' \mid ({P_i}^{s} \cap M_r = \emptyset) \land (c_i \notin M_r) \}$ 
    \ENDFOR
    \FOR{$g_i \in G''$}
        \STATE $G''' \gets \{ g_i \in G'' \mid \forall p \in {P_i}^{s}, \, |d(p) - d(c_i)| \leq T_d \}$
    \ENDFOR
    \FOR{$g_i \in G'''$}
        \STATE $g^* \gets \mathop{\arg\min}\limits_{g_i \in G'''} d(c_i)$
    \ENDFOR\\
    // Optimal Grasp Refinement: Adjust the width and center of $g^*$ to get the final grasp $g^*_f$.\\
    \STATE $R_{ec} \gets f(g^*, M_r)$\\
    \STATE $\left( w'_s, c'_i \right) \gets \left(min{R_{ec}}(w) + e_c, {R_{ec}}(c) \right)$\\
    \STATE $g^*_f \gets (g^*, (w'_s, c'_i))$\\
    \RETURN $g^*_f$
\end{algorithmic}
\end{algorithm}

\subsubsection{Grasp Angle Calibration}
Inappropriate selection of the grasp angle $\theta$ may cause the object to slip or fall during grasping due to the uneven force distribution on both fingers of the parallel-jaw gripper. Therefore, all grasp candidates (denoted by $G$) need to be ensured to have an optimal angle before sampling, as it directly determines the subsequent sampling effectiveness. Here, we use the grasping model in \cite{c20} to obtain the grasp candidates. Specifically, we first extract the edges of the instance mask. For each grasp candidate, we rotate them clockwise in 2-degree intervals until they reach 360 degrees. For each rotation $r$, we find four intersection points between the two long sides of the grasp candidate and the edges, that is, $p_{t_l}$, $p_{b_l}$, $p_{t_r}$, and $p_{b_r}$. 

Subsequently, we calculate the angle $\theta'$ between the vector $\mathbf{v}_{p_l}$ determined by $p_{b_l}$ and $p_{t_l}$ and the vector $\mathbf{v}_{g_u}$ of the long upper side of this grasp candidate, and similarly to get the angle $\theta''$ between the vector $\mathbf{v}_{p_r}$ determined by $p_{b_r}$ and $p_{t_r}$ and the vector $\mathbf{v}_{g_u}$, shown in Fig. \ref{fig4} (b). By subtracting 90 degrees from each of these angles, taking the absolute value, and summing them, we obtain the angle difference for each rotation. Finally, we select the rotation with the smallest angle difference ${R}^*$ given by Eq. \ref{eq:6} and use this rotated box as the new grasp candidate.

\begin{equation} \label{eq:6}
\begin{aligned}
& {R}^* = \underset{R} {\arg\min} \left( \left|\theta'({R}) - 90\right| + \left|\theta''({R}) - 90\right| \right) \\
& \text { s.t. } {R} \in \{0^\circ, 2^\circ, 4^\circ, \ldots, 360^\circ\} 
\end{aligned}
\end{equation}

\subsubsection{Monozone Grasp Analysis}
After grasp angle calibration, we individually analyze each grasp candidate $g_i$ to determine the optimal grasp. Let $G'$ denote the grasp candidate sets after angle calibration. We first examine the relationship between each grasp candidate $g_i$ and the instance mask $M_r$ by checking whether the pixels ${P_i}^{s}$ along the two short sides of the grasp candidate fall within the instance mask. We filter out $g_i$ if any pixel $p$ in ${P_i}^{s}$ is within mask $M_r$. Additionally, to ensure that grasp candidates are concentrated within $M_r$, we also filter out $g_i$ whose centers $c_i$ are not within $M_r$ and get grasp candidate sets $G''$, which is shown in Eq. \ref{eq:7}. 

However, since these processes probably filter all $g_i$ out, we adaptively rotate the viewpoint clockwise to alter the grasp candidate. That is, if no $g_i$ are available from the current viewpoint, we rotate the image 30 degrees at a time and repeat it until the available grasp candidate is found. Since the camera does not rotate and is constrained by hand-eye calibration, we need to project the rotated candidate grasp \(g_i'\) back to the original viewpoint. Here, the parameters \(w'\) and \(h'\) of \(g_i'\) remain unchanged. The angle \(\theta'\) can be adjusted by adding the rotation angle \(\theta_c\) and restricting it to the range \([- \pi/2, \pi/2]\) to convert it back to the angle \(\theta\) of the candidate grasp \(g_i\) under the original viewpoint. For the center \(c_i'(x', y')\) of \(g_i'\), assuming the center of rotation is \(c_r(x_r, y_r)\), the projection relationship from \(c_i'(x', y')\) to the center \(c_i(x, y)\) of the grasp candidate \(g_i\) under the original viewpoint can be obtained from Eq. \ref{eq:8}. The visualization of viewpoint rotation is shown in Fig. \ref{fig5}.

\begin{equation} \label{eq:7}
G'' = \{ g_i \in G' \mid ({P_i}^{s} \cap M_r = \emptyset) \land (c_i \notin M_r) \}
\end{equation}

\begin{equation} \label{eq:8}
\begin{bmatrix}
x \\
y
\end{bmatrix}
=
\begin{bmatrix}
\cos(\theta_c) & -\sin(\theta_c) \\
\sin(\theta_c) & \cos(\theta_c)
\end{bmatrix}
\begin{bmatrix}
x' - x_r \\
y' - y_r
\end{bmatrix}
+
\begin{bmatrix}
x_r \\
y_r
\end{bmatrix}
\end{equation}

Next, based on $G''$, we analyze their relationship with adjacent objects by setting a depth threshold $T_d$, that is, if the depth difference between any $p$ and $c_i$ exceeds $T_d$, the grasp candidate $g_i$ will be filtered out and get grasp candidate sets $G'''$, as shown in Eq. \ref{eq:9}. The above steps can filter out most grasp candidates $g_i$ that could lead the robot to collide with a target object or its adjacent objects, as well as the table.

\begin{figure}[!t]
\vspace{0.2\baselineskip}
\centerline{\includegraphics[width=\columnwidth]{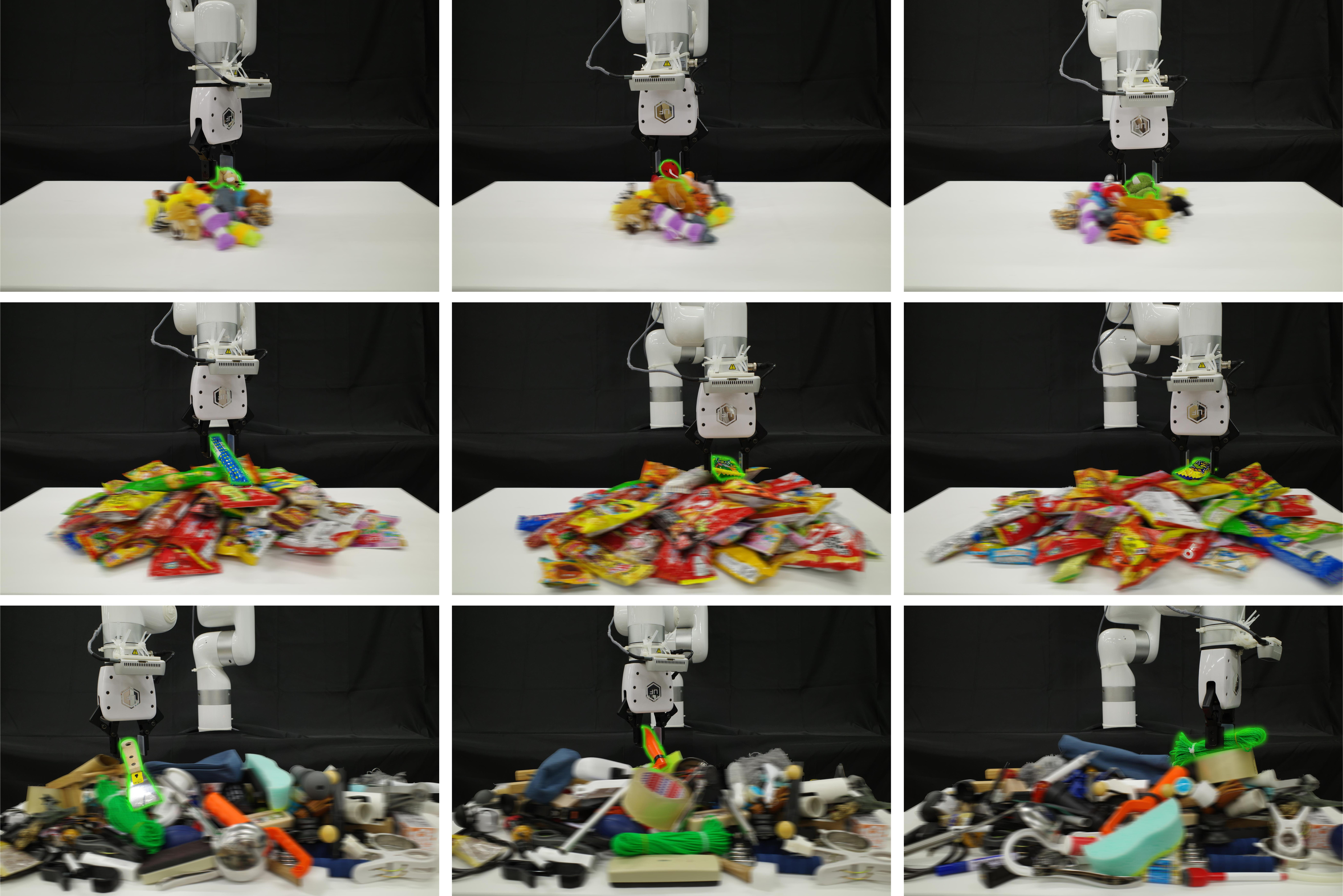}}
\caption{Visualization of OCID grasping dataset. Each subfigure shows different piled objects in different backgrounds.}
\label{fig6}
\end{figure}

\begin{figure}[!t]
\vspace{0.2\baselineskip}
\centerline{\includegraphics[width=\columnwidth]{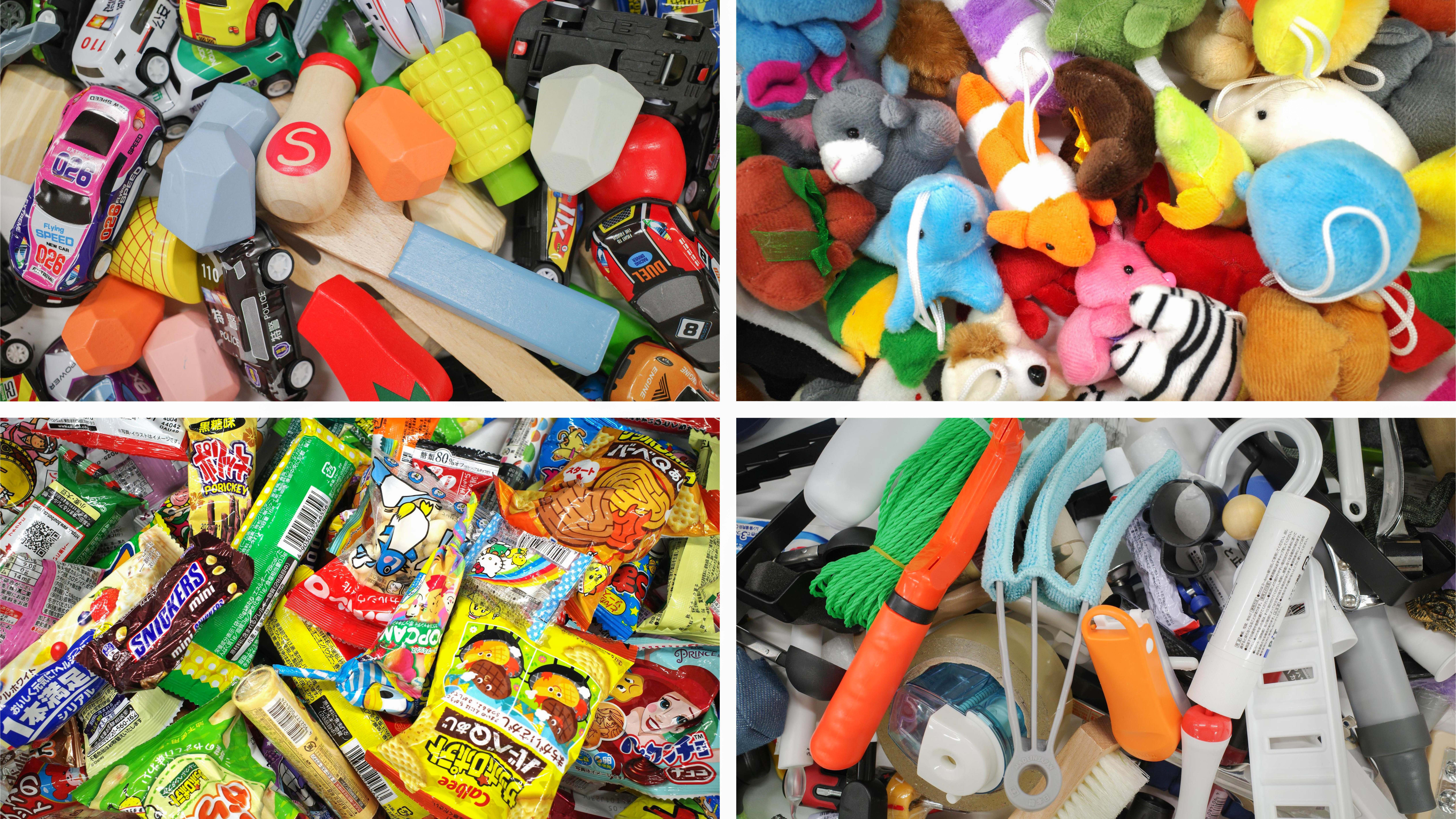}}
\caption{Objects for the grasping experiment: toys, ragdolls, household goods, and snacks (clockwise from top left).}
\label{fig7}
\end{figure}

Finally, based on the sampled grasp candidate sets $G'''$, we use our previous method \cite{c32} to select the $g_i$ with the smallest center pixel depth value $d(c_i)$ as the optimal grasp $g^*$, which is shown in Eq. \ref{eq:10}. This operation ensures a more stable and safer grasp, especially when dealing with objects with complex shapes.

\begin{equation} \label{eq:9}
G''' = \{ g_i \in G'' \mid \forall p \in {P_i}^{s}, \, |d(p) - d(c_i)| \leq T_d \}
\end{equation}

\begin{equation} \label{eq:10}
g^* = \mathop{\arg\min}\limits_{g_i \in G'''} d(c_i)
\end{equation}

\subsubsection{Optimal Grasp Refinement}
Although the optimal grasp $g^*$ was obtained through the previous coarse-to-fine analysis, this analysis will be influenced by random errors from the depth camera, where $g^*$  might still result in collisions with adjacent objects during grasping execution. One way to get around this problem was reported in \cite{c33}, where a series of intervals was defined within the grasp box and the grasp width and position were adjusted based on the relationships between these intervals. However, this method relies on local regions of the depth image with errors, and it is computationally cumbersome. Therefore, we find the minimum rectangle $R_{ec}$ intersecting the optimal grasp $g^*$ and the instance mask $M_r$, and this process can be represented by $f(g^*, M_r)$. Then calculate the shortest width $w_s$ and a new center point $c'_i$ of this grasp through ${min{R_{ec}}(w)}$ and ${R_{ec}}(c)$. Additionally, to mitigate the impact of hand-eye calibration errors, we further expand $w_s$ to $w'_s$ by adding some of the hand-eye calibration translation errors $e_c$ in the $X$ and $Y$-axis. Finally, we use $w'_s$ and $c'_i$ as the new width and center of the grasp for optimal grasp, and denote the final grasp by $g^*_f$, as shown in Eq. \ref{eq:11}, Eq. \ref{eq:12}, and Algorithm 2.

\begin{equation} \label{eq:11}
R_{ec} = f(g^*, M_r)
\end{equation}

\begin{equation} \label{eq:12}
\left( w'_s, c'_i \right) = \left(min{R_{ec}}(w) + e_c, {R_{ec}}(c) \right)
\end{equation}

\section{Experiments}
In this section, we validate the effectiveness of our method by conducting benchmarking studies. Firstly, we compare it with baseline grasping methods in various mid-clutter (up to 20 objects) and high-clutter (up to 50 objects) scenes. Then we increase the number of objects to 100 (extreme-clutter) and analyze the effectiveness of PSP and MSP.

\subsection{Experimental Settings}
\subsubsection{Setting for Grasping Model}
The baseline methods are categorized into two groups. The first group includes GGCNN \cite{c17}, GGCNN2 \cite{c18}, GRconvnet \cite{c20}, SEnet \cite{c19}, and FCGnet \cite{c21}, which are suitable for mid-clutter scenarios. The second group comprises DexNet 4.0 \cite{c4} and GraspNet \cite{c5}, which are tailored for high-clutter scenarios. 

For the first group, since the pre-trained models were all trained on the Cornell Grasping Dataset \cite{c15} (only one object in each fixed white background), their performance in cluttered environments is limited. Therefore, we merge the OCID Grasping Dataset \cite{c11, c34} (with different piled objects, backgrounds, sensor-to-scene distance, viewpoint angle, and lighting conditions, as shown in Fig. \ref{fig6}.) into the Cornell Grasping Dataset and retrain these models using the parameter settings specified in their original papers (except that all uses the RGB-D modality). Specifically, these models were trained on a single NVIDIA RTX 4090 GPU with 24 GB of memory. The system is Ubuntu 22.04, and the deep learning framework is PyTorch 2.3.1 with CUDA 12.1. Before training, we randomly shuffle the entire dataset, using 90\% for training and 10\% for testing. During training, the data are uniformly cropped to fit the acceptable sizes, the number of training epochs is set to 50, and data augmentation (random zoom and random rotation) is applied. For testing, we use the same metric \cite{c18} to report the detection accuracy (Acc) of these methods. According to this metric, a grasp is considered valid when it satisfies two conditions: the Intersection over Union (IoU) score between the ground truth and predicted grasp rectangles is over 25\%, and the offset between the orientation of the ground truth rectangle and that of the predicted grasp rectangle is less than \SI{30}{\degree}.

For the second group, we directly use their pre-trained models: the parallel-jaw version of DexNet 4.0 and the planar version of GraspNet \cite{c5, c35}. Finally, unless specified, the segmentation and grasp candidate prediction components of PMSGP use the pre-trained models of SAM and GRconvnet in all experiments.

\begin{figure}[!t]
\vspace{-0.2\baselineskip}
\centerline{\includegraphics[width=\columnwidth]{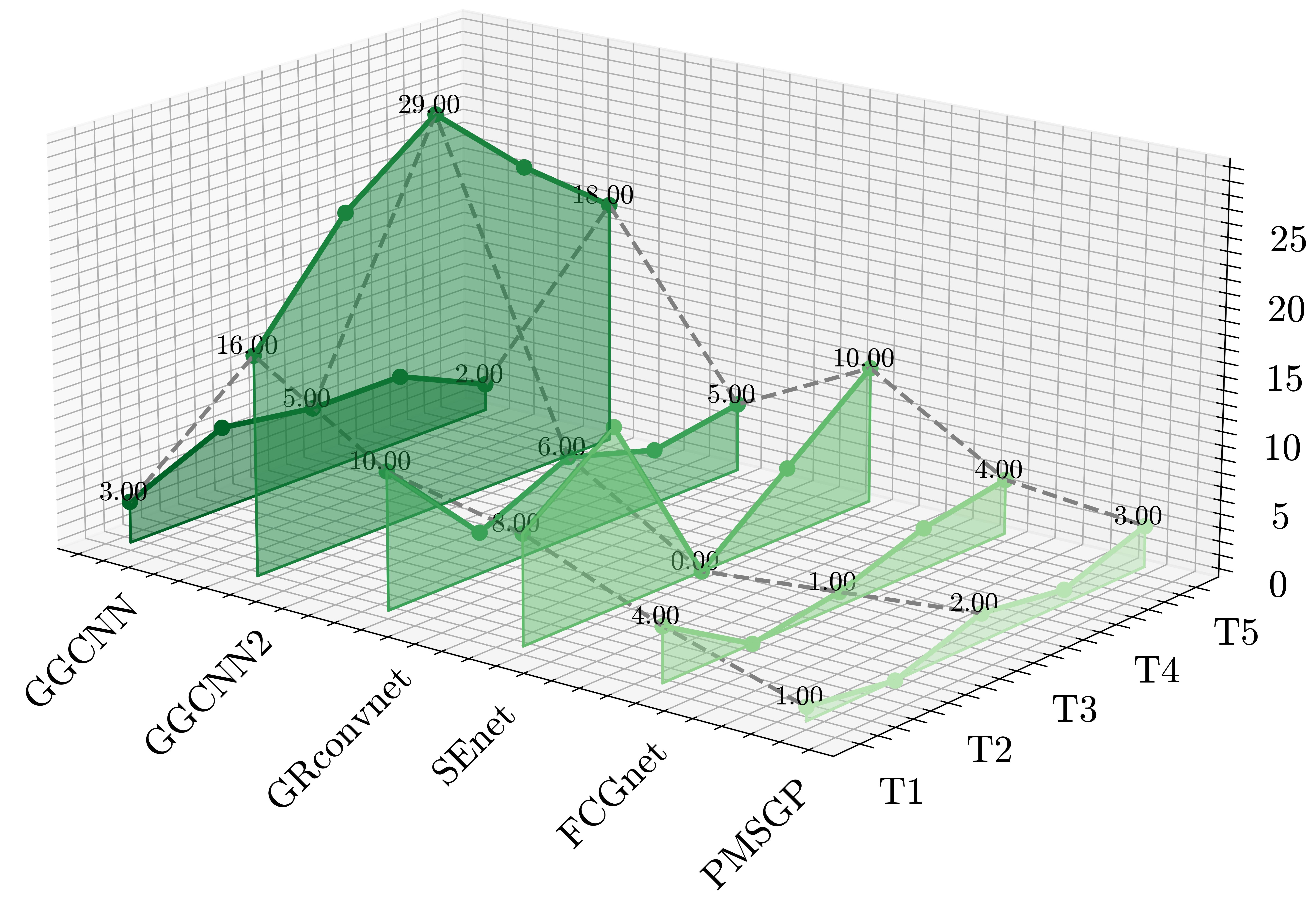}}
\caption{Line graph showing GSR of PMSGP and first-group baselines. The horizontal axis represents different methods 
and the depth axis represents trials from T1 to T5. The vertical axis represents the number of grasp failures. We emphasize the number of grasp failures (T1, T3, T5) in each method with dots, and connect them with dashed lines to better show the difference.}
\label{fig8}
\end{figure}

\begin{table}[!t]
\caption{GSR comparison among PMSGP and first-group baselines\label{tab:table1}}
\centering
\begin{tabular}{cccccccc}
\toprule 
Methods & T1 & T2 & T3 & T4 & T5 & Acc (\%) & GSR (\%)\\
\midrule
GGCNN & 3 & 6 & 5 & 5 & 2 & 22.3 & 82.6\\
GGCNN2 & 16 & 24 & 29 & 23 & 18 & 37.7 & 47.6\\
GRconvnet & 10 & 3 & 6 & 4 & 5 & 52.0 & 78.1\\
SEnet & 8 & 13 & 0 & 5 & 10 & 45.0 & 73.5\\
FCGnet & 4 & 0 & 1 & 3 & 4 & 52.0 & 89.3\\
PMSGP & 1 & 0 & 2 & 1 & 3 & - & \textbf{93.5}\\
\bottomrule
\end{tabular}
\end{table}

\begin{figure*}[!t]
\centerline{\includegraphics[width=\textwidth]{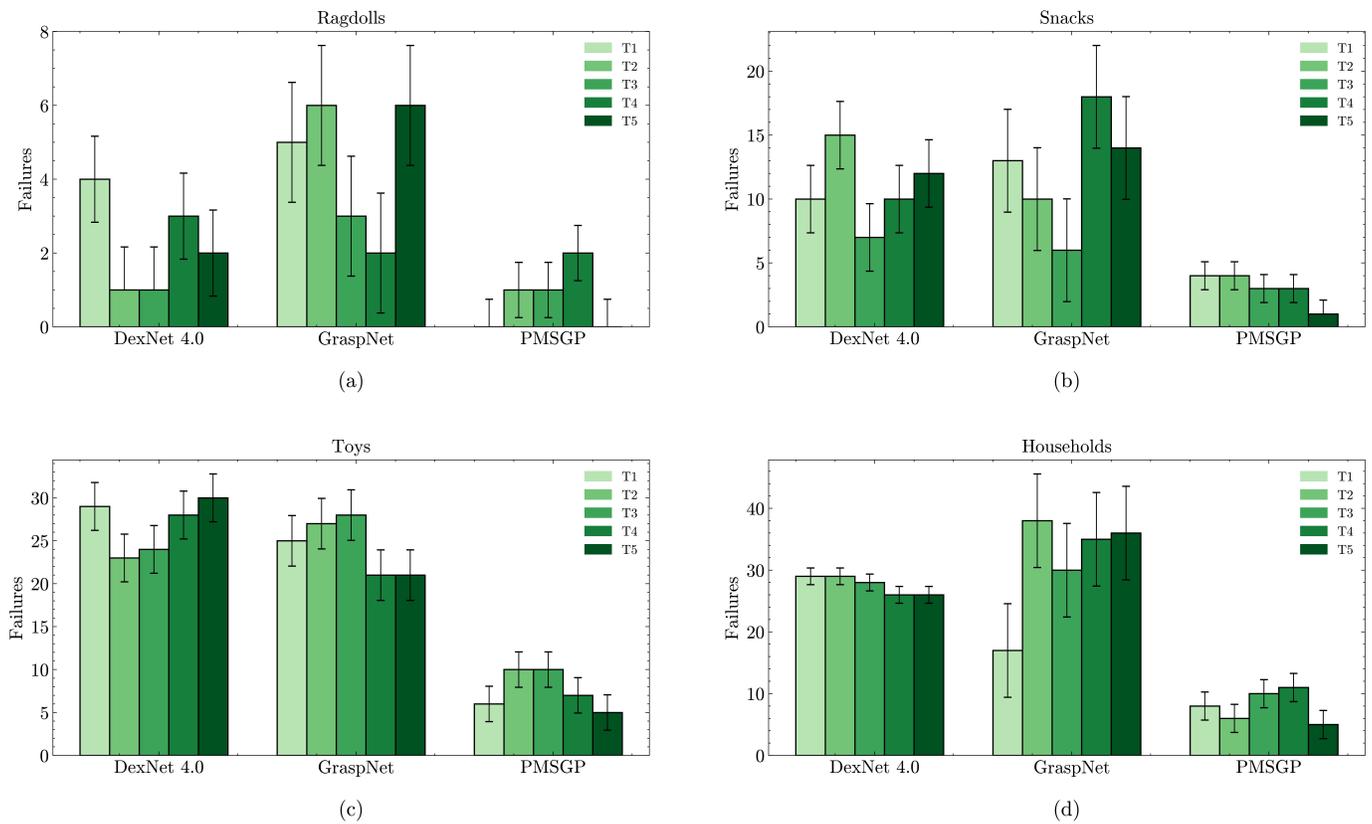}}
\caption{Bar graphs showing GSR of PMSGP and second-group baselines. (a), (b), (c), and (d) represent the results of testing ragdolls, snacks, toys, and household goods. In each subfigure, the vertical axis represents the number of grasp failures, and the horizontal axis represents different methods with five trials. 
We show the positive and negative errors at the top of each bar by calculating the mean of the number of grasp failures across all trials for each method.}
\label{fig9}
\end{figure*}

\begin{table*}[!t]
\caption{GSR comparison among PMSGP and second-group baselines\label{tab:table2}}
\centering
\resizebox{\textwidth}{!}{
\begin{tabular}{c|cccccc|cccccc|cccccc|cccccc}
\toprule 
\multirow{2}{*}{Methods} & \multicolumn{6}{c|}{Ragdolls} & \multicolumn{6}{c|}{Snacks}  & \multicolumn{6}{c|}{Toys} & \multicolumn{6}{c}{Household goods}\\
\cmidrule(lr){2-7}
\cmidrule(lr){8-13}
\cmidrule(lr){14-19}
\cmidrule(lr){20-25}
& T1 & T2 & T3 & T4 & T5 & GSR (\%) & T1 & T2 & T3 & T4 & T5 & GSR (\%) & T1 & T2 & T3 & T4 & T5 & GSR (\%) & T1 & T2 & T3 & T4 & T5 & GSR (\%)\\
\midrule
DexNet 4.0 & 4 & 1 & 1 & 3 & 2 & 95.8 & 10 & 15 & 7 & 10 & 12 & 82.2 & 29 & 23 & 24 & 28 & 30 & 65.1 & 29 & 29 & 28 & 26 & 26 & 64.4\\
GraspNet & 5 & 6 & 3 & 2 & 6 & 92.0 & 13 & 10 & 6 & 18 & 14 & 80.4 & 25 & 27 & 28 & 21 & 21 & 67.2 & 17 & 38 & 30 & 35 & 36 & 61.6\\
PMSGP & 0 & 1 & 1 & 2 & 0 & \textbf{98.4} & 4 & 4 & 3 & 3 & 1 & \textbf{94.3} & 6 & 10 & 10 & 7 & 5 & \textbf{86.8} & 8 & 6 & 10 & 11 & 5 & \textbf{86.2}\\
\bottomrule
\end{tabular}}
\end{table*}

\begin{figure*}[!t]
\centerline{\includegraphics[width=\textwidth]{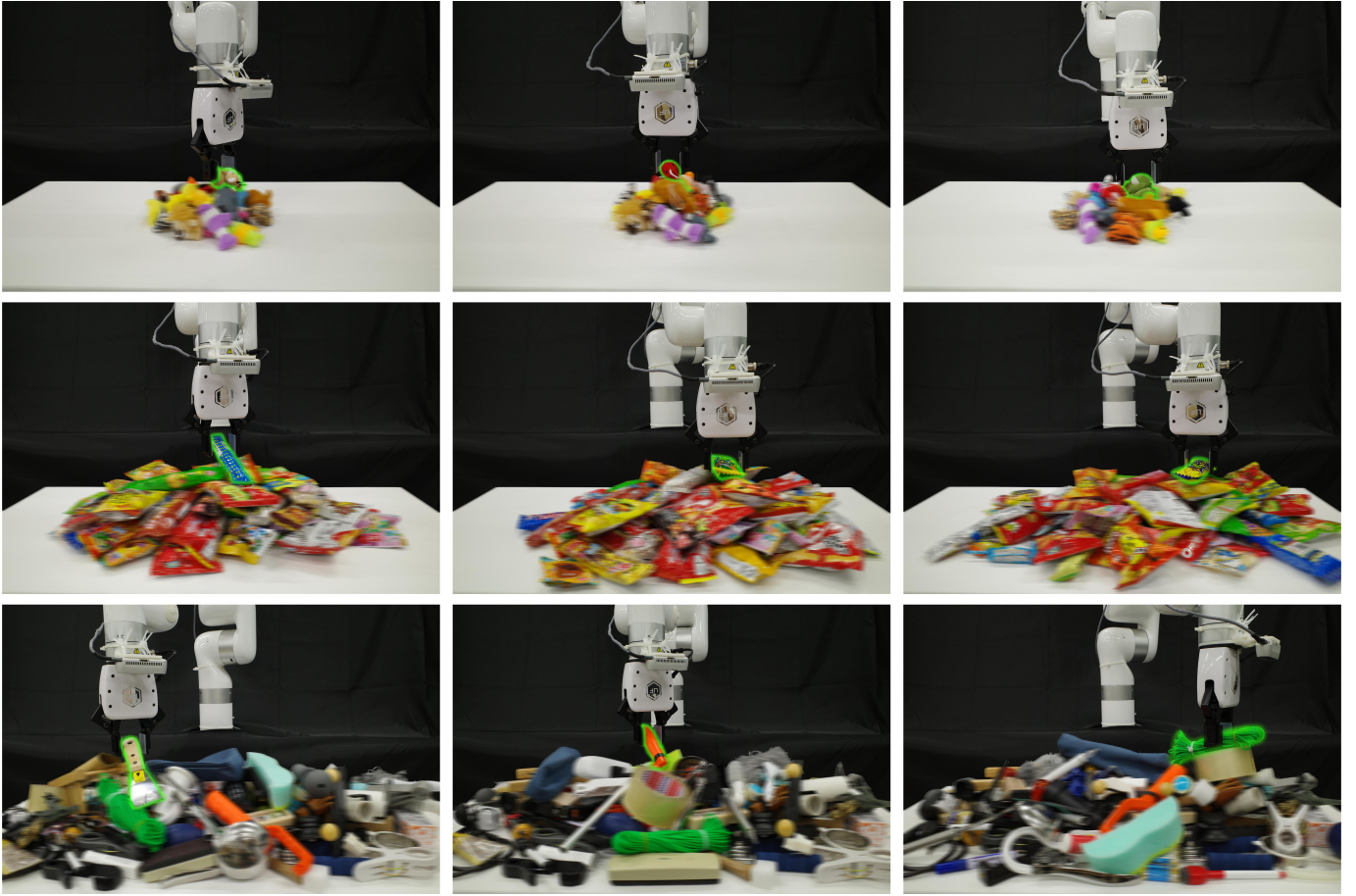}}
\caption{Performance of PMSGP on objects of different grasping difficulties and scene complexities. The first row is a mid-clutter scene consisting of 20 ragdolls, the second row is a high-clutter scene consisting of 50 snacks, and the last row is an extreme-clutter scene consisting of 100 household goods. The topmost object in each subfigure is emphasized by green lines.}
\label{fig10}
\end{figure*}

\subsubsection{Setting for Real Grasping}
Our grasping system consists primarily of an Intel RealSense D435 depth camera and a UFactory xArm 5 robot. We employ an eye-in-hand architecture, with the camera mounted on the robot’s distal end and facing downward. The objects used in our grasping experiments are divided into four categories, with a total of 300 novel objects: 50 ragdolls (Category 1), 100 snacks (Category 2), 50 toys (Category 3), and 100 household goods (Category 4), respectively, as shown in Fig. \ref{fig7}. Category 1 is the easiest and as the category level goes up, the grasping difficulty rises, too. 

Before grasping, we fill the depth hole and set a depth value threshold (with the upper limit of 40 $cm$ and the lower limit of 10 $cm$ explained earlier in Section IV-A) to ensure that the grasp is executed within this range \cite{c17}. During grasping, each method is tested in five trials per experiment, and the number of failed grasps in each trial (T) is recorded. The grasp success rate (GSR) is calculated by dividing the total number of successful grasps by the total number of grasp attempts across five trials. In addition, to improve experimental safety and ensure all objects are grasped in each trial, we provide minimal manual assistance during the experiments. Specifically, if an object fails to be grasped 2-3 times, we manually pick up the object and count it as a failure. Additionally, if an object moves out of the camera view, it is repositioned with manual intervention. Similarly, if a grasped object moves out of the robot's workspace, causing it to stop, the object is repositioned manually, too. 

\subsection{Comparision Studies}
\subsubsection{Comparision with Baseline Methods in Mid-clutter}
We compare PMSGP with the baseline methods in the first group. We used 10 snacks and 10 household goods to form a mid-clutter scene. The results are shown in Table \ref{tab:table1}. PMSGP achieves a GSR of 93.5\% (100/107), with only 7 grasp failures, which is far superior to other baselines except for FCGnet. Additionally, we found that some baselines perform well on the dataset but not in real grasping. For example, GGCNN2 has a GSR of only 47.6\% (100/210) with a total of 110 grasp failures, indicating that this method does not generalize well to novel objects in mid-clutter. Finally, we also visualize the result in Fig. \ref{fig8} to better show the gap between PMSGP with other baseline methods. As shown in this figure, it is very clear that our method's number of failures is much smaller with little variance across the trials.

\subsubsection{Comparision with Baseline Methods in High-clutter}
DexNet 4.0 and GraspNet are considered state-of-the-art for learning-based 4-DOF and 6-DOF grasping, respectively. Therefore, to demonstrate the effectiveness of PMSGP's grasping capability, we compare it with these two methods. It is important to note that we just compare the parallel gripper version of DexNet 4.0 and the planar version of GraspNet. Moreover, we conducted experiments in high-clutter scenes composed of 50 ragdolls, 50 snacks, 50 toys, and 50 household goods, respectively. The experimental results are shown in Table \ref{tab:table2}, indicating that PMSGP achieves GSR of 98.4\% (250/254) for ragdolls, 94.3\% (250/265) for snacks, 86.8\% (250/288) for toys, and 86.2\% (250/290) for household goods. All surpassed DexNet 4.0 and GraspNet. More importantly, as the difficulty in grasping increases, the gap between PMSGP and the baseline methods becomes more obvious. For example, when grasping toys and household goods, PMSGP's GSR exceeds theirs by up to 20\%, demonstrating the high reliability of our method. We also visualize these results in Fig. \ref{fig9}. It is obvious that the bar length of our method is shorter than that of other methods in each subfigure, and it varies little across the trials and has smaller errors.

\subsection{Ablation Studies}
\subsubsection{Effectiveness of Pyramid Sequencing Policy}
To emphasize the importance of TVA in PSP, we decompose PSP into TVA and Cross-prompted CPS for separate validation. For TVA, we use a mid-clutter scenario consisting of 20 household goods, because PMSGP without TVA can only grasp within a very limited area. Additionally, PMSGP without TVA uses the best quality pixel as the prompt for segmentation, with other aspects consistent with PMSGP. For CPS, we use an extreme-clutter scenario consisting of 100 snacks, as their complex appearance helps fully validate the effectiveness of CPS. Here, PMSGP without CPS uses the first segmentation, with other aspects also consistent with PMSGP. The experimental results are shown in Table \ref{tab:table3} and \ref{tab:table4}. The GSR of PMSGP without TVA is only 74.6\% (100/134), whereas the original PMSGP achieves 90\% (100/111), showing a substantial gap of 15.4\%. It is therefore deemed crucial to select the right order of grasping targets to avoid occlusions. Similarly, the GSR of PMSGP without CPS is 79.1\% (500/632), compared with the original PMSGP's 90\% (500/564), proving the effectiveness of CPS. In addition, we show some grasping performance of PMSGP on objects of different grasping difficulties at different complexity in Fig. \ref{fig10}. It can be seen that PMSGP is always able to avoid most occlusions and target the topmost objects.

\subsubsection{Effectiveness of Monozone Sampling Policy}
In this section, we demonstrate the effectiveness of MSP under the highest grasping difficulty, specifically in extremely cluttered scenes composed of 100 household goods. These objects exhibit the greatest variation in materials, shapes, and appearances compared to other objects that we use. Additionally, the version without MSP uses quality-based sampling, while other aspects remain consistent with PMSGP. The experimental results, as shown in Table \ref{tab:table5}, indicate that the GSR of PMSGP without MSP is 75\% (500/667), compared to 84.9\% (500/589) for PMSGP. Two cases differ by approximately 10\%, illustrating the obvious advantage of MSP over quality-based sampling in high-difficulty scenarios.

\begin{table}[!t]
\caption{Impact of with or without TVA\label{tab:table3}}
\centering
\begin{tabular}{ccccccc}
\toprule 
Methods & T1 & T2 & T3 & T4 & T5 & GSR (\%)\\
\midrule
Without TVA & 6 & 5 & 10 & 6 & 7 & 74.6 \\
PMSGP & 4 & 1 & 1 & 4 & 1 & \textbf{90.0} \\
\bottomrule
\end{tabular}
\end{table}

\begin{table}[!t]
\caption{Impact of with or without CPS\label{tab:table4}}
\centering
\begin{tabular}{ccccccc}
\toprule 
Methods & T1 & T2 & T3 & T4 & T5 & GSR (\%)\\
\midrule
Without CPS & 23 & 29 & 20 & 28 & 32 & 79.1 \\
PMSGP & 19 & 14 & 9 & 9 & 13 & \textbf{88.7} \\
\bottomrule
\end{tabular}
\end{table}

\begin{table}[!t]
\caption{Impact of with or without MSP\label{tab:table5}}
\centering
\begin{tabular}{ccccccc}
\toprule 
Methods & T1 & T2 & T3 & T4 & T5 & GSR (\%)\\
\midrule
Without MSP & 25 & 35 & 32 & 34 & 41 & 75.0 \\
PMSGP & 22 & 14 & 17 & 12 & 24 & \textbf{84.9} \\
\bottomrule
\end{tabular}
\end{table}

\subsection{Failure Case Analysis}
In the above experiments, we performed more than 7,000 grasp attempts and achieved a total of 5,800 successful grasps. More importantly, we tested PMSGP's capability in extreme-clutter scenarios involving 100 novel objects, and the GSR is stable between 85\% to 89\%. To the best of our knowledge, no previous work has demonstrated similar performance. However, PMSGP still encounters some failures. The first issue is slippage during grasp execution, which occurs due to the smooth surface of the object. To address this, we plan to use parallel jaw grippers with high-friction finger pads or wrap the fingers with textured tape.
Furthermore, errors from the depth camera can cause some grasps to deviate slightly from the topmost object. This problem can be mitigated by using a high-precision industrial depth camera. While our method reduces most occlusions during grasping, in certain cases, the lower part of the topmost object may still be occluded by the one beneath it. To optimize this, we are going to use amodal instance segmentation \cite{c36} to predict the occluded parts of the object, thereby getting the complete mask of the object. Finally, when objects with similar depths are tightly packed together, they will be difficult to grasp. This challenge can be overcome by designing a deformable parallel-jaw gripper that adapts under force to separate the objects, like \cite{c37}.

\section{Conclusion}
In this paper, we proposed the Pyramid-Monozone Synergistic Grasping Policy (PMSGP), which first employed the Pyramid Sequencing Policy (PSP) to determine the optimal grasping sequence and then applied the Monozone Sampling Policy (MSP) to identify the best grasp candidate. PMSGP enabled the robot to effectively mitigate occlusions in densely cluttered environments, significantly reducing the difficulty of grasping and enhancing the grasping success rate. We conducted over 7,000 real-world grasping attempts on 300 novel objects across various levels of clutter, including mid-clutter (20 objects), high-clutter (50 objects), and extreme-clutter (100 objects), demonstrating that PMSGP significantly outperforms seven competitive grasping methods.

Future work can be divided into two major parts. The first part can focus on addressing the issues highlighted in the Failure Case Analysis to enhance the method proposed in this paper. The second part can involve using this method as a baseline and extending it to human-robot interaction for specific object retrieval. For instance, safely grasping a specific object in a cluttered scene without interfering with other objects, and securely handing it over to a person should be noteworthy.

\phantomsection


\addcontentsline{toc}{section}{References} 

\begin{thebibliography}{99}
\bibitem{c1} R. M. Murray, Z. Li, and S. S. Sastry, \emph{A Mathematical Introduction to Robotic Manipulation}. Boca Raton, FL, USA: CRC Press, 2017.

\bibitem{c2} D. Prattichizzo and J. C. Trinkle, “Grasping,” in \emph{Springer Handbook of Robotics}, Berlin, Germany: Springer 2008.

\bibitem{c3} B. Kehoe, A. Matsukawa, S. Candido, J. Kuffner, and K. Goldberg, “Cloud-based robot grasping with the google object recognition engine,” in \emph{Proc. IEEE Int. Conf. Robot. Automat.}, 2013, pp. 4263–4270.

\bibitem{c4} J. Mahler et al., “Learning ambidextrous robot grasping policies,” \emph{Sci. Robot.}, vol. 4, no. 26, pp. 1–12, 2019.

\bibitem{c5} H. S. Fang, M. Gou, C. Wang, and C. Lu, “Robust grasping across diverse sensor qualities: The GraspNet-1Billion dataset,” \emph{Int. J. Robot. Res.}, vol. 42, no. 12, pp. 1094–1103, 2023.

\bibitem{c6} A. Vaswani, N. Shazeer, N. Parmar, J. Uszkoreit, L. Jones, A.N. Gomez, Ł. Kaiser, and I. Polosukhin, “Attention is all you need,” in \emph{Proc. Conf. Neural Informat. Process. Syst.}, 2017, pp. 6000–6010.

\bibitem{c7} S. Hochreiter, and J. Schmidhuber, "Long short-term memory," \emph{Neural Comput.}, vol. 9, no. 8, pp. 1735–1780, 1997.

\bibitem{c8} T. Brown, b. Mann, N. Ryder, M. Subbiah, J.D. Kaplan, P. Dhariwal, A. Neelakantan, P. Shyam, G. Sastry, A. Askell, and S. Agarwal, "Language models are few-shot learners," in \emph{Proc. Conf. Neural Informat. Process. Syst.}, 2020, pp. 1877–1901.

\bibitem{c9} A. Dosovitskiy, L. Beyer, A. Kolesnikov, D. Weissenborn, X. Zhai, T. Unterthiner, M. Dehghani, M. Minderer, G. Heigold, S. Gelly, and J. Uszkoreit, "An image is worth 16x16 words: Transformers for image recognition at scale," 2020, \emph{arXiv:}2010.11929.

\bibitem{c10} R. Girshick, J. Donahue, T. Darrell, and J. Malik, "Rich feature hierarchies for accurate object detection and semantic segmentation," in \emph{Proc. IEEE Conf. Comput. Vis. Pattern Recognit.}, 2014, pp. 580-587.

\bibitem{c11} S. Ainetter and F. Fraundorfer, “End-to-end trainable deep neural network for robotic grasp detection and semantic segmentation from RGB,” in \emph{Proc. IEEE Int. Conf. Robot. Automat.}, 2021, pp. 13452–13458.

\bibitem{c12} J. Li and D. J. Cappelleri, “Sim-Suction: Learning a suction grasp policy for cluttered environments using a synthetic benchmark,” \emph{IEEE Trans. Robot.}, vol. 40, pp. 316–331, 2024.

\bibitem{c13} C. Rosales, J. M. Porta, and L. Ros, “Grasp optimization under specific contact constraints,” \emph{IEEE Trans. Robot.}, vol. 29, no. 3, pp. 746–757, 2013.

\bibitem{c14} F. T. Pokorny, K. Hang, and D. Kragic, “Grasp moduli spaces,” in \emph{Proc. Robot.: Sci. Syst.}, 2013.

\bibitem{c15} I. Lenz, H. Lee, and A. Saxena, “Deep learning for detecting robotic grasps,” \emph{Int. J. Robot. Res.}, vol. 34, no. 4–5, pp. 705–724, 2015.

\bibitem{c16} K. He, X. Zhang, S. Ren, and J. Sun, "Deep residual learning for image recognition," in \emph{Proc. IEEE Conf. Comput. Vis. Pattern Recognit.}, 2016, pp. 770-778.

\bibitem{c17} D. Morrison, P. Corke, and J. Leitner, “Closing the loop for robotic grasping: A real-time, generative grasp synthesis approach,” in \emph{Proc. Robot.: Sci. Syst.}, 2018.

\bibitem{c18} D. Morrison, P. Corke, and J. Leitner, “Learning robust, real-time, reactive robotic grasping,” \emph{Int. J. Robot. Res.}, vol. 39, no. 2-3, pp. 183–201, 2020.

\bibitem{c19} S. Yu, D.-H. Zhai, Y. Xia, H. Wu, and J. Liao, “SE-ResUNet: A novel robotic grasp detection method,” \emph{IEEE Robot. Automat. Lett.}, vol. 7, no. 2, pp. 5238–5245, 2022.

\bibitem{c20} S. Kumra, S. Joshi, and F. Sahin, “Antipodal robotic grasping using
generative residual convolutional neural network,” in \emph{Proc. IEEE/RSJ Int. Conf. Intell. Robots Syst.}, 2020, pp. 9626–9633.

\bibitem{c21} M. Shan, J. Zhang, H. Zhu, C. Li, and F. Tian, "Grasp Detection Algorithm Based on CPS-ResNet," in \emph{Proc. IEEE Int. Conf. Image Process. Comput. Vis. Mach. Learn.}, 2022, pp. 501-506.

\bibitem{c22} H. Cao, G. Chen, Z. Li, Q. Feng, J. Lin, and A. Knoll, “Efficient grasp detection network with Gaussian-based grasp representation for robotic manipulation,”\emph{IEEE/ASME Trans. Mech.}, vol. 28, no. 3, pp. 1384–1394, 2022.

\bibitem{c23} S. Yu, D.-H. Zhai, and Y. Xia, “CGNet: Robotic grasp detection in heavily cluttered scenes,” \emph{IEEE/ASME Trans. Mech.}, vol. 28, no. 2, pp. 884–894, 2023.

\bibitem{c24} J. Mahler et al., “Dex-Net 1.0: A cloud-based network of 3D objects for robust grasp planning using a multi-armed bandit model with correlated rewards,” in \emph{Proc. IEEE Int. Conf. Robot. Automat.}, 2016, pp. 1957–1964.

\bibitem{c25} J. Mahler et al., “Dex-Net 2.0: Deep learning to plan robust grasps with synthetic point clouds and analytic grasp metrics,” in \emph{Proc. Robot.: Sci. Syst.}, 2017.

\bibitem{c26} J. Mahler, M. Matl, X. Liu, A. Li, D. Gealy, and K. Goldberg, “Dex-Net 3.0: Computing robust vacuum suction grasp targets in point clouds using a new analytic model and deep learning,” in \emph{Proc. IEEE Int. Conf. Robot. Automat.}, 2018, pp. 5620–5627.

\bibitem{c27} J. Mahler and K. Goldberg, “Learning deep policies for robot bin picking by simulating robust grasping sequences,” in \emph{Conf. Robot Learn.}, 2017, pp. 515–524.

\bibitem{c28} H. S. Fang, C. Wang, M. Gou, and C. Lu, “GraspNet-1billion: A large scale benchmark for general object grasping,” in \emph{Proc. IEEE Conf. Comput. Vis. Pattern Recognit.}, 2020, pp. 11444–11453.

\bibitem{c29} H. S. Fang et al., “AnyGrasp: Robust and efficient grasp perception in spatial and temporal domains,” \emph{IEEE Trans. Robot.}, vol. 39, no. 5, pp. 3929–3945, 2023.

\bibitem{c30} A. Kirillov et al., “Segment anything,” 2023, \emph{arXiv:} 2304.02643.

\bibitem{c31} N. Kanopoulos, N. Vasanthavada, and R. L. Baker, “Design of an image edge detection filter using the Sobel operator,” \emph{IEEE J. Solid-State Circuits.}, vol. 23, no. 2, pp. 358–367, 1988.

\bibitem{c32} C. Li, P. Zhou, N. Y. Chong, "Safety-optimized Strategy for Grasp Detection in High-clutter Scenarios,". in \emph{Proc. Int. Conf. Ubiquitous Robots}, 2024, pp. 501-506.

\bibitem{c33} P. Raj, A. Kumar, V. Sanap, T. Sandhan, and L. Behera, “Towards object agnostic and robust 4-DoF table-top grasping,” in \emph{Proc. IEEE Int. Conf. Autom. Sci. Eng.}, 2022, pp. 963–970.

\bibitem{c34} M. Suchi, T. Patten, and M. Vincze, “EasyLabel: A semi-automatic pixelwise object annotation tool for creating robotic RGB-D datasets,” in \emph{Proc. IEEE Int. Conf. Robot. Automat.}, 2019, pp. 6678–6684.

\bibitem{c35} F.-J. Chu, R. Xu, and P. A. Vela, “Real-world multiobject, multigrasp detection,” \emph{IEEE Robot. Automat. Lett.}, vol. 3, no. 4, pp. 3355–3362, 2018.

\bibitem{c36} J. Zhang, Y. Gu, J. Gao, H. Lin, Q. Sun, X. Sun, X. Xue, and Y. Fu, "LAC-Net: Linear-Fusion Attention-Guided Convolutional Network for Accurate Robotic Grasping Under the Occlusion." 2024, \emph{arXiv:} 2408.03238.

\bibitem{c37} Zeng et al., “Robotic pick-and-place of novel objects in clutter with multi-affordance grasping and cross-domain image matching,” \emph{Int. J. Robot. Res.}, vol. 41, no. 7, pp. 690–705, 2022.

\end{thebibliography}
\end{document}